\documentclass[11pt,a4paper]{article}
\usepackage[hyperref]{acl2021}
\usepackage{times}
\usepackage{latexsym}
\usepackage{graphicx}
\usepackage{algorithm}
\usepackage{algorithmicx}
\usepackage{algpseudocode}

\renewcommand{\algorithmiccomment}[1]{\bgroup\hfill//~#1\egroup}
\usepackage{amsmath,amssymb}

\usepackage{multirow}
\usepackage{booktabs}
\usepackage{subfigure}
\usepackage{microtype}

\aclfinalcopy % Uncomment this line for the final submission
\def\aclpaperid{3433} %  Enter the acl Paper ID here
\title{Discontinuous Named Entity Recognition as Maximal Clique Discovery}

\author{
  Yucheng Wang\textsuperscript{\thanks{ \quad The two authors contribute equally.} }, 
  Bowen Yu\textsuperscript{\footnotemark[1]},
  Hongsong Zhu\\ 
  \bf 
  Tingwen Liu\textsuperscript{\thanks{\quad Corresponding author.}}, 
  Nan Yu, 
  Limin Sun\\
  Institute of Information Engineering, Chinese Academy of Sciences \\
 School of Cyber Security, University of Chinese Academy of Sciences \\
  {\tt \{wangyucheng,yubowen,zhuhongsong\}@iie.ac.cn} \\
  {\tt \{liutingwen,yunan,sunlimin\}@iie.ac.cn}
}
\date{}
\begin{document}
\maketitle
\begin{abstract}
% Named entity recognition (NER) remains challenging when entity mentions can be discontinuous. Existing methods break the recognition process into several sequential steps.  In training, they predict conditioned on the golden intermediate results, while at inference relying on the model prediction of the previous steps, which introduces exposure bias. To solve this problem, we reformulate the discontinuous NER into a task of discovering maximal cliques in segment graphs. The nodes and edges are predicted in a single stage without exposure bias by a deep model. In the graph, a node denotes a span (a continuous entity on its own, or a part of discontinuous entities), and an edge links two nodes that belong to the same entity. By finding maximal cliques and concatenating the spans in each clique, we can yield predicted entity mentions. Experiments on three benchmark datasets show that our method works very well for both discontinuous and conventional NER and achieved SoTA performance on each dataset, with F1 score gains up to 3.5 percentage points.
Named entity recognition (NER) remains challenging when entity mentions can be discontinuous. Existing methods break the recognition process into several sequential steps.  
In training, they predict conditioned on the golden intermediate results, while at inference relying on the model output of the previous steps, which introduces exposure bias. 
To solve this problem, we first construct a segment graph for each sentence, in which each node denotes a segment (a continuous entity on its own, or a part of discontinuous entities), and an edge links two nodes that belong to the same entity. 
The nodes and edges can be generated respectively in one stage with a grid tagging scheme and learned jointly using a novel architecture named Mac.
Then discontinuous NER can be reformulated as a non-parametric process of discovering maximal cliques in the graph and concatenating the spans in each clique.
Experiments on three benchmarks show that our method outperforms the state-of-the-art (SOTA) results, with up to 3.5 percentage points improvement on F1, and achieves 5x speedup over the SOTA model.\footnote{The source code is available at \url{https://github.com/131250208/InfExtraction}}
% Experiments on three benchmark datasets show that our method works very well for both discontinuous and conventional NER, and achieved SoTA performance on each dataset, with F1 score gains up to 3.5 percentage points.
\end{abstract}

\section{Introduction}
Named Entity Recognition (NER) is the task of detecting mentions of real-world entities from text and classifying them into predefined types. 
NER benefits many natural language processing applications (e.g., information retrieval~\cite{berger2017information}, relation extraction~\cite{yu2019beyond}, and question answering~\cite{khalid2008impact}).

% This is normally cast as a sequence label- ing problem where each token is assigned a label that represents its entity type.

\begin{figure}
    \centering
    \includegraphics[width=\columnwidth]{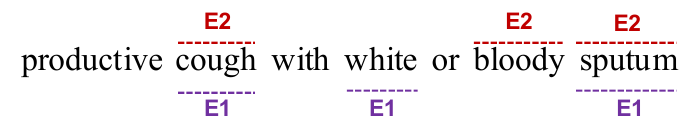}
    \caption{An example involving discontinuous mentions. Entities are highlighted with colored underlines.}
    \label{fig:example}
    \vspace{-0.1in}
\end{figure}

NER methods have been extensively investigated and researchers have proposed effective ones.
Most prior approaches~\cite{huang2015bidirectional, chiu2016named, gridach2017character, zhang2018chinese, gui2019cnn,mengge2020coarse} cast this task as a sequence labeling problem where each token is assigned a label that represents its entity type.
Their underlying assumption is that an entity mention should be a short span of text~\cite{muis2016learning}, and should not overlap with each other.
While such assumption is valid for most cases, it does not always hold, especially in clinical corpus~\cite{pradhan2015evaluating}.
%First, entities with discontinuous structures are very common in clinical corpus~\cite{pradhan2015evaluating}. 
%Second, entities can be overlapping, where some tokens may belong to multiple mentions~\cite{kim2003genia}.
For example, Figure~\ref{fig:example} shows two discontiguous entity mentions with overlapping segments.
Thus, there is a need to move beyond continuous entities and devise methods to extract discontinuous ones. 

%Therefore, it is critical to consider nested mentions for real-world applications and downstream tasks.

%Towards this goal, there are broadly two kinds of efforts: combination-based and transition-based.
Towards this goal, current state-of-the-art (SOTA) models can be categorized into two classes: combination-based and transition-based.
Combination-based models first detect all the overlapping spans and then learn to combine these segments with a separate classifier~\cite{wang2019combining};
Transition-based models incrementally label the discontinuous spans through a sequence of shift-reduce actions~\cite{dai-etal-2020-effective}.
Although these methods have achieved reasonable performance, they continue to have difficulty with the same problem: \textbf{exposure bias}~\cite{zhang2019bridging}.
Specifically, combination-based methods use the gold segments to guide the classifier during the training process while at inference the input segments are given by a trained model, leading to a gap between training and inference~\cite{wang2019combining}.
For transition-based models, at training time, the current action relies on the golden previous actions, while in the testing phase, the entire action sequence is generated by the model.
As a result, a skewed prediction will further deviate the predictions of the follow-up actions. 
Such accumulated discrepancy may hurt the performance.

In order to overcome the limitation of such prior works, we propose Mac, a \underline{\textbf{Ma}}ximal \underline{\textbf{c}}lique discovery based discontinuous NER model.
The core insight behind Mac is that all (potentially discontinuous) entity mentions in the sentence can naturally form a segment graph by interpreting their contained continuous segments as nodes, and connecting segments of the same entity to each other as edges.
Then the discontinuous NER task is equivalent to finding the maximal cliques from the graph, which is a well-studied problem in graph theory.
So, the question that remains is how to construct such a segment graph.
We decompose it into two uncoupled subtasks, segment extraction (SE) and edge prediction (EP) in Mac.
Typically, given an $n$-token sentence, two $n \times n$ tag tables are formed for SE and EP respectively where each entry captures the interaction between two individual tokens.
SE is then regarded as a labeling problem where tags are assigned to distinguish the boundary tokens of each segment, which have benefits in identifying overlapping segments.
EP is converted as the problem of aligning the boundary tokens of segments contained in the same entity.
Overall, the tag tables of SE and EP are generated independently, and will be consumed together by a maximum clique searching algorithm to recover desired entities from them, thus immune from the exposure bias problem.

%spans that either form entities on their own or present as parts of a discontiguous entity can naturally form a segment graph

%In this paper, we present a one-stage method named xxxx, which is capable of exactly identifying discontiguous entities while immune from the exposure bias problem. 
% 我们的核心动机是可以自然地构成一个graph，在图上识别不连贯实体就是在查找极大团，而这是一个graph领域被广泛研究的问题
% 那么剩下的挑战就是如何识别segment并且确定segment之间的联系
% 我们提出了一个novel 的table标注方案，包含segment

% , including CADEC, ShARe 13 and ShARe 14 datasets
We conducted experiments on three standard discontinuous NER benchmarks. 
Experiments show that Mac can effectively recognize discontinuous entity mentions without sacrificing the accuracy on continuous mentions.
This leads to a new state-of-the-art (SOTA) on this task, with substantial gains of up to 3.5\% absolute percentage points over previous best reported result.
Lastly, we show that in the runtime experiments on GPU environments, Mac is about five times faster than the SOTA model.
\section{Related Work}
% Our work is inspired by three lines of research: discontinuous NER, joint extraction, and maximal clique discovery.

\textbf{Discontinuous NER} requires to identify all entity mentions that have discontinuous structures.
To achieve this end, several researchers introduced new position indicators into the traditional BIO tagging scheme so that the sequential labeling models can be employed~\cite{tang2013recognizing,metke2016concept,dai2017medication,tang2018recognizing}.
However, this model suffers from the label ambiguity problem due to the limited flexibility of the extended tag set.
As the improvement, \citeauthor{muis2016learning}~\shortcite{muis2016learning} used hyper-graphs to represent entity spans and their combinations, but did not completely resolve the ambiguity issue~\cite{dai-etal-2020-effective}. 
\citeauthor{wang2019combining}~\shortcite{wang2019combining} presented a pipeline framework which first detects all the candidate spans of entities and then merges them into entities.
By decomposing the task into two inter-dependency steps, this approach does not have the ambiguity issue, but meanwhile being susceptible to exposure bias.
Recently, \citeauthor{dai-etal-2020-effective}~\shortcite{dai-etal-2020-effective} constructed a transition action sequence for recognizing discontinuous and overlapping structure.
At training time, it predicts with the ground truth previous actions as condition while at inference it has to select the current action based on the results of previous steps, leading to exposure bias. 
In this paper, for the first time we propose a one-stage method to address discontinuous NER while without suffering from the ambiguity issue, realizing the consistency of training and inference.
%Different from previous methods, our model unambiguously extract discontinuous segments and merge them together without any coupled steps, realizing the consistency of training and inference.

\textbf{Joint extraction} aims to detect entity pairs along with their relations using a single model~\cite{yu2019joint}.
Discontinuous NER is related to joint extraction where the discontiguous entities can be viewed as relation links between segments~\cite{wang2019combining}.
Our model is motivated by TPLinker~\cite{wang2020tplinker}, which formulates joint extraction as a token pair linking problem by aligning the boundary tokens of entity pairs.
% We extend the concepts from refining networks to tackle conflict among words.
The main differences between our model and TPLinker are two-fold:
(1) We propose a tailor-designed tagging scheme for recognizing discontinuous segments;
(2) The maximal clique discovery algorithm is introduced into our model to accurately merge the discontinuous segments.

\textbf{Maximal clique discovery} is to find a clique of maximum size in a given graph~\cite{dutta2019finding}. 
Here, a clique is a subset of the vertices all of which are pairwise adjacent.
Maximal clique discovery finds extensive application across diverse domains~\cite{stix2004finding,boginski2005statistical,imbiriba2017band}.
In this paper, we reformulate discontinuous NER as the task of maximal clique discovery by constructing a segment graph and leveraging the classic B-K backtracking algorithm~\cite{bron1973algorithm} to find all the maximum cliques as the entities.

\section{Methodology}

\begin{figure}
    \centering
    \includegraphics[width=0.95\columnwidth]{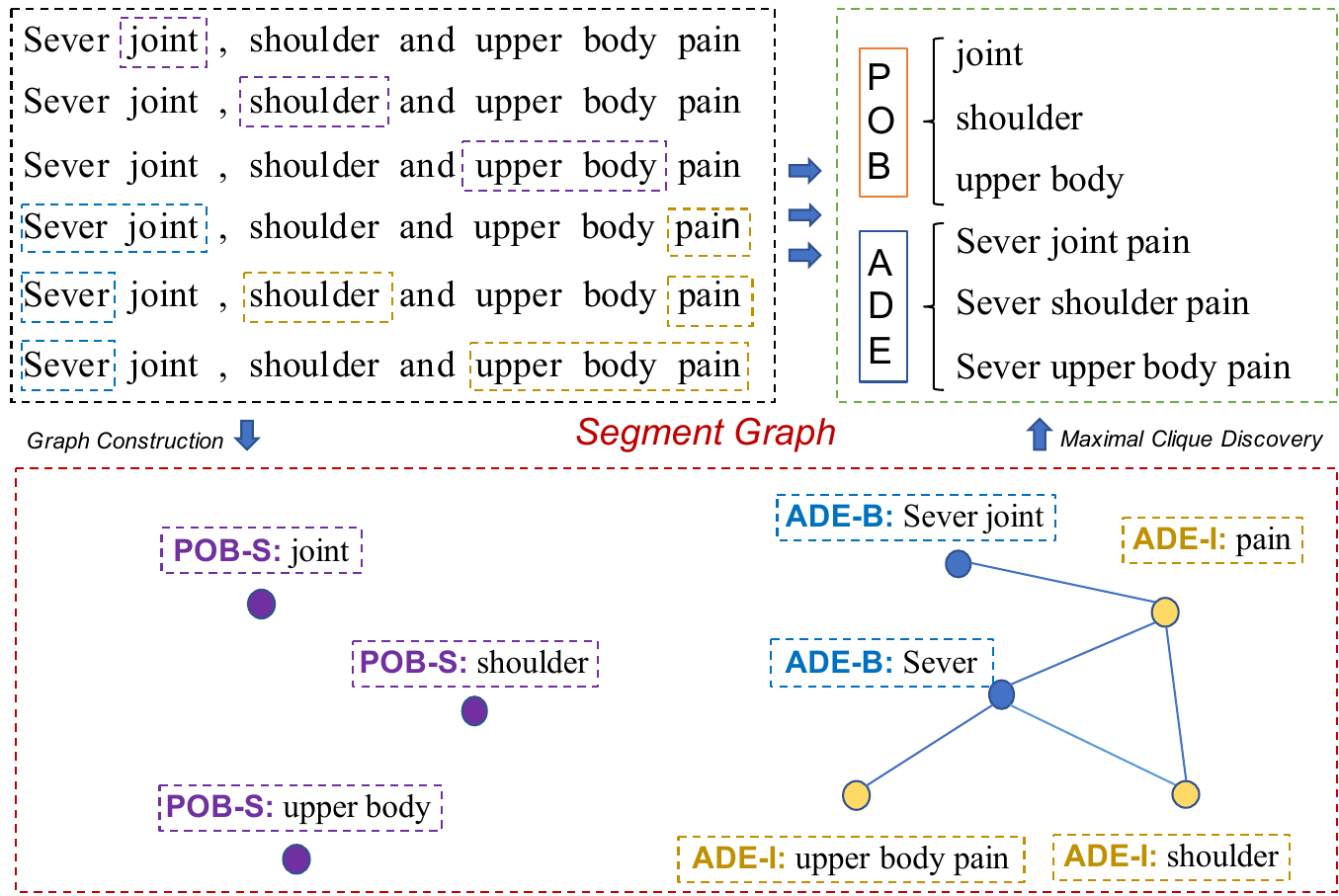}
    \caption{An example of the extraction process.}
    \label{fig:extraction}
    \vspace{-0.1in}
\end{figure}

In graph theory, a clique is a vertex subset of an undirected graph where every two vertices in the clique are adjacent, while a maximal clique is the one that cannot be extended by including one more adjacent vertex. That means each vertex in the maximal clique has close relations with each other, and no other vertex can be added, which is similar to the relations between segments in a discontinuous entity. 
Based on this insight, we claim that discontinuous NER can be equivalently interpreted as discovering maximal cliques from a segment graph, where nodes represent segments that either form entities on their own or present as parts of a discontinuous entity, and edges connect segments that belong to the same entity mention.
% A maximal clique is a clique that cannot be extended by including one more adjacent vertex, that is.
% A maximal clique is a clique that cannot be extended by including one more adjacent vertex, that is, a clique which does not exist exclusively within the vertex set of a larger clique. Some authors define cliques in a way that requires them to be maximal, and use other terminology for complete subgraphs that are not maximal.

Considering the maximum clique searching process is usually non-parametric~\cite{bron1973algorithm}, discontinuous NER is actually decomposed into two subtasks: segment extraction and edge prediction, to respectively create the nodes and edges of the segment graph.
Their prediction results can be generated independently with our proposed grid tagging scheme, and will be consumed together to construct a segment graph, so that the maximal clique discovery algorithm can be applied to recover desired entities. 
The overall extraction process is depicted in Figure~\ref{fig:extraction}.
Next, we will first introduce our grid tagging scheme and its decoding workflow. 
Then we will detail the Mac, a \underline{\textbf{Ma}}ximal \underline{\textbf{c}}lique discovery based discontinuous NER model based on this tagging scheme.

\subsection{Grid Tagging Scheme}
Inspired by \newcite{wang2020tplinker}, we implement single-stage segment extraction and edge prediction based on a novel grid tagging scheme.
Given an $n$-token sentence, our scheme constructs an $n \times n$ tag table by enumerating all possible token pairs and giving each token pair the tag(s) based on their relation(s).
Note that one token pair may have multiple tags according to the pre-defined tag set.

\begin{figure}
    \centering
    \includegraphics[width=0.95\columnwidth]{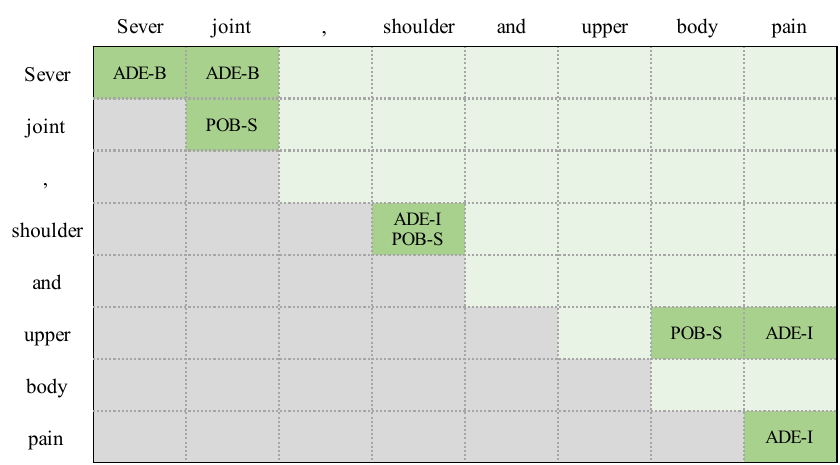}
    \caption{A tagging example for segment extraction.}
    \label{fig:segment}
    \vspace{-0.1in}
\end{figure}

\subsubsection{Segment Extraction}
As demonstrated in Figure~\ref{fig:example}, entity mentions could overlap with each other. 
To make our model capable of extracting such overlapping segments, we construct a two-dimensional tag table.
Figure~\ref{fig:segment} provides an example.
A pair of tokens $(t_i, t_j)$ will be assigned with a set of labels if a segment from $t_i$ to $t_j$ belongs to the corresponding categories.
Considering $j \geq i$, we discard the lower triangle region of the tag table, so $\frac{n^2+n}{2}$ grids are actually generated for an $n$-token sentence.
In practice, the BIS tagging scheme is adopted to represent if a segment is a continuous entity mention (X-S) or locates at the beginning (X-B) or inside (X-I) of a discontinuous entity of type X.
For example, (\textit{upper}, \textit{body}) is assigned with the tag POB-S since ``\textit{upper body}'' is a continuous entity of type Part of Body (POB).
 And the tag of (\textit{Sever}, \textit{joint}) is ADE-B as ``\textit{Sever joint}'' is a beginning segment of the discontinuous mention ``\textit{Sever joint pain}'' of type Adverse Drug Event (ADE).
Meanwhile, ``\textit{joint}'' is also recognized as an entity since there is a POB-S tag in the place of (\textit{joint}, \textit{joint}), thus the overlapping segment extraction problem is solved.

\subsubsection{Edge Prediction}

\begin{figure}
    \centering
    \includegraphics[width=0.95\columnwidth]{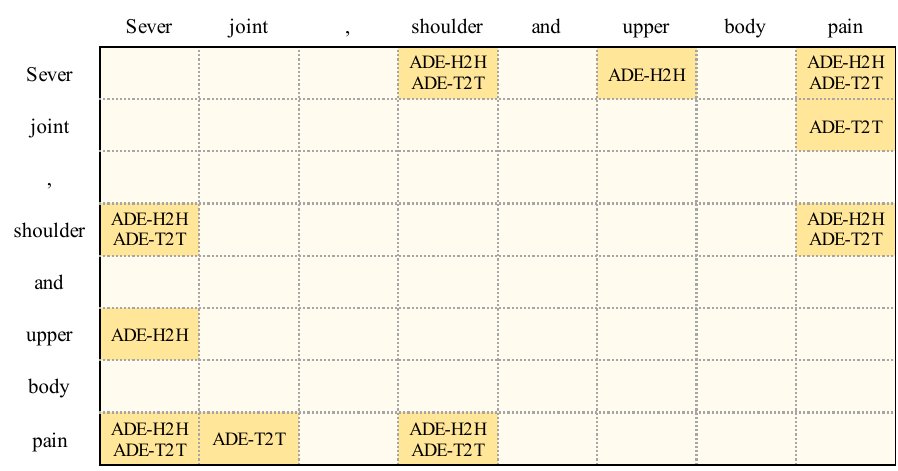}
    \caption{A tagging example for edge prediction.}
    \label{fig:edge}
    \vspace{-0.1in}
\end{figure}

Edge prediction is to construct the links between segments of the same entity mention by aligning their boundary tokens.
The tagging scheme is defined as follows: (1) head to head (X-H2H) indicates it locates in a place ($t_i$, $t_j$) where $t_i$ and $t_j$ are respectively the beginning tokens of two segments which constitute the same entity of type X; (2) tail to tail (X-T2T) is similar to X-H2H, but focusing on the ending token.
As shown in Figure~\ref{fig:edge}, ``\textit{Sever}'' has the ADE-H2H and ADE-T2T relations to ``\textit{shoulder}'' and ``\textit{pain}'', because the type of the discontinuous entity mention ``\textit{Sever shoulder pain}'' is Adverse Drug Event .
The same logic goes for other tags in the matrix.

\subsection{Decoding Workflow}
Formally, the decoding procedure is summarized in Algorithm~\ref{alg:decoding}. 
The segment tagging table $\rm S$ and edge tagging table $\rm E$ of a sentence $\rm T$ serve as the inputs. 
Firstly, we extract all the typed segments through decoding $\rm S$.
Then we construct a segment graph $G$, in which segments that belong to the same entity (decoded from $\rm E$) have edges with each other.
Figure~\ref{fig:extraction} gives an example. 
%In this case, we can find two kinds of maximal cliques: 1) \textit{single-vertex}: the purple vertex; 2) \textit{multiple-vertex}: multiple vertices linked to each other.
Correspondingly, we can yield a continuous entity mention from the single-vertex clique directly, and concatenate segments in each multiple-vertex clique following their original sequential order in $\rm T$ to recover discontinuous entity mentions. 
%Then we discover all the maximal cliques and recover the discontinuous entity mentions by concatenating the segments in each clique following their original sequential order in $T$.
We choose the classic B-K backtracking algorithm~~\cite{bron1973algorithm} for finding the maximal cliques in $G$, which takes $O(3^{\frac{m}{3}})$ time, where $m$ is the number of nodes.

%\begin{figure}
%    \centering
%    \includegraphics[width=0.95\columnwidth]{Figures/edge_decoding.pdf}
%    \caption{An example of the extraction process.}
%    \label{fig:edge_decoding}
%    \vspace{-0.1in}
%\end{figure}

\begin{algorithm}[t]
% \scriptsize
\footnotesize
% \tiny
% \small
\caption{Decoding Procedure}
\label{alg:decoding}
\begin{algorithmic}[1]
\Require The segment tagging results $\rm S$ and edge tagging results $\rm E$ of sentence $\rm T$. ${\rm S}(t_i,t_j)$ and ${\rm E}(t_i,t_j)$ respectively denote the tag set of token pair $(t_i,t_j)$ in two schemes.
\Ensure ${\rm R}=\{(e_k, t_k)\}_{k=1}^m$, $e_k$, $t_k$ are respectively the text and the type of the $k$-th entity.

\State  Initialize the edge set $\rm{A}$ and entity set $\rm{R}$ with $\varnothing$
\State  Obtain the segment set $\rm{N}$ by decoding $S$.
    \For{segment $s \in { \rm N}$}
        \For{segment $g \in {\rm N}$}
        	\State  Define type $\leftarrow$ the entity type of $s$ or $g$   			
                \If {type-H2H $\in$ E(s.start, g.start) $\And$ type-T2T $\in$ E(s.end, g.end)}
                    \State Add (s, g) to $A$
                \EndIf
        \EndFor
    \EndFor
    \State Construct the segment graph $G$ based on N and A
    \State Find the maximal cliques $C$ in $G$ with the B-K algorithm
    \For{clique $c \in C$}
        \State  Define $t \leftarrow$ the entity type of a random segment in $c$
        \State Concat the segments of $c$ with their order in $\rm T$ as $e$
        \State Add $(e, t)$ to ${\rm R}$
\EndFor

\State \Return R
\end{algorithmic}
\end{algorithm}

\begin{algorithm}[t]
% \scriptsize
\footnotesize
% \tiny
% \small
\caption{B-K Backtracking Algorithm}
\label{alg:b_k}
\begin{algorithmic}[1]
\Require The graph $G$ 
\Ensure the set of all maximal cliques: $\rm{C}$.
\State Initialize $\rm{C}$ and two vertex sets R, X with $\varnothing$
\State Define  P $ \leftarrow$ the node set of $G$ 

\Function {BronKer}{R, P, X}
    \If{P $ = \varnothing~\And$ X$= \varnothing$}
        \State Add R to $\rm{C}$
    \EndIf
    \For{$v \in $ P }
    	\State Define N$(v) \leftarrow$ the neighbor set of $v$
        \State \Call{BronKer}{ R $\cup$ N$(v)$, P $\cap$ N$(v)$, X $\cap$ N$(v)$}
        \State P $\leftarrow$ P $\backslash$ ${v}$
        \State X $\leftarrow$ X $\cup$ ${v}$
    \EndFor
\EndFunction

\State \Call{BronKer}{R, P, X} \Comment{call the BronKer function}

\State \Return $\rm{C}$
\end{algorithmic}
\end{algorithm}

\subsection{Model Structure}
\begin{figure}
    \centering
    \includegraphics[width=0.95\columnwidth]{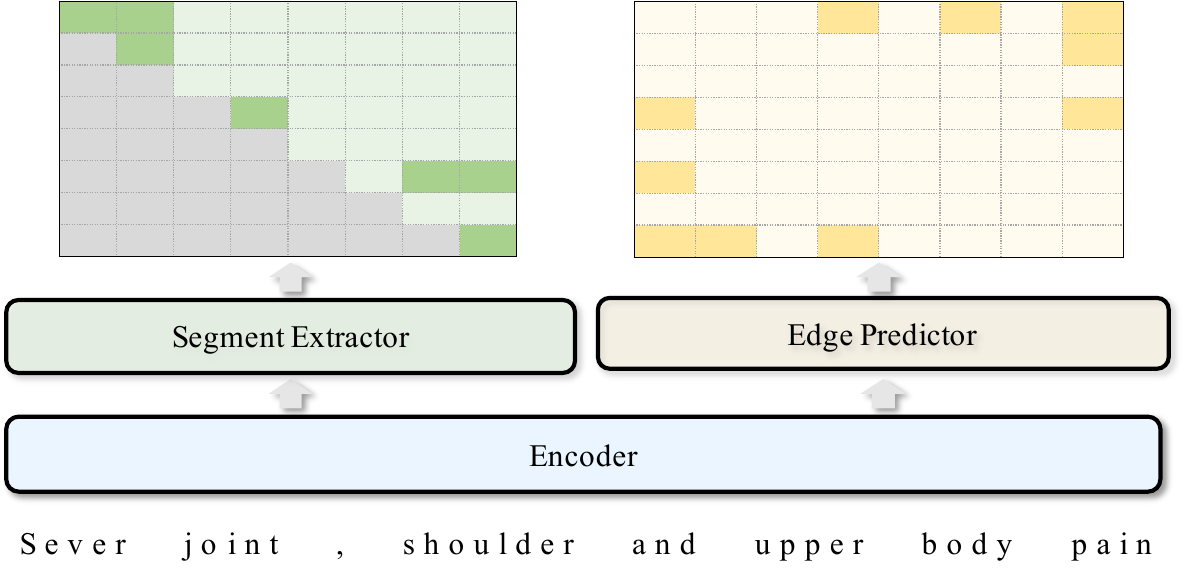}
    \caption{The overall structure of the Mac model.}
    \label{fig:model}
    \vspace{-0.1in}
\end{figure}

With the grid tagging scheme, we propose an end-to-end neural architecture named Mac.
%, which first encodes the sentence into contextual representations. 
%Then, two types of grid encoders are built For segment extraction and edge prediction respectively. 
Figure~\ref{fig:model} reveals the overview structure.

\subsubsection{Token Representation} 
Given an $n$-token sentence $[t_1,\cdots,t_n]$, we first map each token $t_i$ into a low-dimensional contextual vector $\mathbf{h_i}$ with a basic encoder. Then we generate two representations, $\mathbf{h}^s_{i}$ and $\mathbf{h}^e_{i}$, as the task-specific features for the segment extractor and the edge predictor, respectively:
\begin{align}
    \label{equ:tok_rep1}
    \mathbf{h}^s_i = \mathbf{W}^s_h \cdot \mathbf{h}_i + \mathbf{b}^s_h, \\
    \label{equ:tok_rep2}
    \mathbf{h}^e_i = \mathbf{W}^e_h \cdot \mathbf{h}_i + \mathbf{b}^e_h,
\end{align}
where $\mathbf{W}^*_h$ is a parameter matrix and $\mathbf{b}^*_h$ is a bias vector to be learned during training.

% >>>>>>>>>>>>>>>>>> Segment Extractor
\subsubsection{Segment Extractor}

The probability that a pair of tokens are the boundary tokens of a segment can be represented as:
\begin{equation}
    \label{equ:bd_toks}   
    P(t_i, t_j) = P(e = t_j|b = t_i)P(b = t_i),
\end{equation}
where $b$ and $e$ denotes the beginning token and ending token. 
In our tagging scheme (Figure~\ref{fig:segment}), we have a fixed beginning token $t_i$ at the $i$-th row, and take the given beginning token as the condition to label the corresponding ending token, so $P(b = t_i)$ in the $i$-th row is always 1. 
Hence, all we need to do is to calculate $P(e = t_j|b = t_i)$.

Inspired by \newcite{conLN} and \citeauthor{yu2021semi}~\shortcite{yu2021semi}, we levderage the Conditional Layer Normalization (CLN) mechanism to model the conditional probability. 
% Since the idea of CLN derives from Conditional Batch Normalization~\cite{de2017modulating}, they share a similar pattern. 
That is, a conditional vector is introduced as extra contextual information to generate the gain parameter $\gamma$ and bias $\lambda$ of the well known layer normalization mechanism~\cite{ba2016layer} as follows:
% to do
% \begin{equation}
\begin{align}
&{\rm CLN}(\mathbf{c}, \mathbf{x}) = \gamma_c \odot (\frac{\mathbf{x}-\mu}{\sigma}) + \lambda_c,  \\
\mu &= \frac{1}{d}\sum_{i=1}^d x_i, \;\;\; \sigma = \sqrt{\frac{1}{d} \sum_{i=1}^d (x_i-\mu)^2}, \\
\gamma_c &= \mathbf{W}_{\alpha} \mathbf{c}+\mathbf{b}_{\alpha}, \;\;\; 
\lambda_c = \mathbf{W}_{\beta} \mathbf{c}+\mathbf{b}_{\beta}.
\label{eq.cln}
\end{align}
% \end{equation}

% \begin{equation}
% {\rm CLN}(\mathbf{c}, \mathbf{x}) = \gamma_c \odot (\frac{\mathbf{x}-\mu}{\sigma}) + \lambda_c, 
% \label{eq.ln}
% \end{equation}
% \begin{equation}
% \mu = \frac{1}{d}\sum_{i=1}^d x_i, \;\;\; \sigma = \sqrt{\frac{1}{d} \sum_{i=1}^d (x_i-\mu)^2}, 
% \label{eq.ln}
% \end{equation}
% \begin{equation}
% \gamma_c = \mathbf{W}_{\alpha} \mathbf{c}+\mathbf{b}_{\alpha}, \;\;\; 
% \lambda_c = \mathbf{W}_{\beta} \mathbf{c}+\mathbf{b}_{\beta}
% \label{eq.cln},
% \end{equation}
where $\mathbf{c}$ and $\mathbf{x}$ are the conditional vector and input vector respectively. 
$x_i$ denotes the $i$-th element of $\mathbf{x}$, $\mu$ and $\sigma$ are the mean and standard deviation taken across the elements of $\mathbf{x}$, respectively. 
$\mathbf{x}$ is firstly normalized by fixing the mean and variance and then scaled and shifted by $\gamma_c$ and $\lambda_c$ respectively.
Based on the CLN mechanism, the representation of token pair $(t_i, t_j)$ being a segment boundary can be defined as:
\begin{equation}
    \label{equ:hsk_seg_cln}
    \mathbf{h}^{sb}_{i,j} = {\rm CLN}(\mathbf{h}^s_i, \mathbf{h}^s_j).
\end{equation}

In this way, For different $t_i$, different LN parameters are generated, which results in effectively adapting $\mathbf{h}_j$ to be more $t_i$-specific.

Furthermore, besides the features of boundary tokens, we also consider inner tokens and segment length to learn a better segment representation. 
Specifically, we deploy a LSTM network~\cite{hochreiter1997long} to compute the hidden states of inner tokens, and use a looking-up table to embed the segment length. 
Since the ending token is always behind the beginning one, in each row $r_i$, only the tokens behind $t_i$ will be fed into the LSTM. 
We take the hidden state outputted at each time step $t_j$ as the inner token representation of the segment $s_{i:j}$.
Then the representation of a segment from $t_i$ to $t_j$ can be defined as follows:
\begin{align}
    \label{equ:hsk_seg_lstm}
    \mathbf{h}^{in}_{i:j} &= {\rm LSTM}(\mathbf{h}^s_i, ..., \mathbf{h}^s_j), j \geq i, \\
    \label{equ:hsk_seg_len_emb}
    \mathbf{e}^{len}_{i:j} &= {\rm Emb}(j - i), j \geq i, \\
    \mathbf{h}^{s}_{i:j} &= \mathbf{h}^{sb}_{i,j} + \mathbf{h}^{in}_{i:j} + \mathbf{e}^{len}_{i:j}.
\end{align}

%Finally, we calculate the tag score vector by:
%\begin{align}
%    \label{equ:classifiers}
%    \mathbf{o}^s_{i:j} = \mathbf{W}^\alpha_o \cdot \mathbf{h}^s_{i:j} + \mathbf{b}^\alpha_o, 
%\end{align}
%where each dimension of $\mathbf{o}^s_{i:j}$ denotes the score of a segment tag between $t_i$ and $t_j$.

% >>>>>>>>>>>>>>>>>>>>>>>>>> Edge Predictor
\subsubsection{Edge Predictor}

Edge prediction is similar with segment extraction since they all need to learn the representation of each token pair.
The key differences are summarized in the following two aspects: (1) the distance between segments is usually not informative, so the length embedding $\mathbf{e}^{len}_{i:j}$ is valueless in edge prediction; 
% the distance between segments is usually ruleless, so the length embedding $\mathbf{e}^{len}_{i:j}$ is useless in edge prediction;
(2) encoding the tokens between segments may carry noisy semantics for correlation tagging and aggravate the burden of training, so no $\mathbf{h}_{i:j}^{in}$ is required.
Under such considerations, we represent each token pair for edge prediction as:
\begin{equation}
    \label{equ:int_rel_cln}
     \mathbf{h}^e_{i,j} = {\rm CLN}(\mathbf{h}^e_i, \mathbf{h}^e_j).
\end{equation}

%Analogously, $(t_i,t_j)$'s tag score vector can be calculated by Equation~\ref{equ:corr_classifier}:
%\begin{align}
%    \label{equ:corr_classifier}
%    \mathbf{o}^e_{i,j} &= \mathbf{W}^e_o \cdot \mathbf{h}^e_{i,j} + \mathbf{b}^e_o.
%\end{align}

\subsection{Training and Inference}

In practical, our grid tagging scheme aims to tag most relevant labels for each token pair, so it can be seen as a multi-label classification problem.
Once having the comprehensive token pair representations ($\mathbf{h}^{s}_{i:j}$ and $\mathbf{h}^{e}_{i:j}$), we can build the multi-label classifier via a fully connected network.
Mathematically, the predicted probability of each tag for $(t_i,t_j)$ can be estimated via:
\begin{equation}
    \label{equ:corr_classifier}
    \mathbf{p}^{\mathcal{I}}_{i,j} = {\rm sigmoid(}\mathbf{W}^{\mathcal{I}} \cdot \mathbf{h}^{\mathcal{I}}_{i,j} + \mathbf{b}^{\mathcal{I}}),
\end{equation}
where $\mathcal{I} \in \{\text{s}, \text{e}\}$ is the symbol of subtask indicator, denoting segment extraction and edge prediction respectively, and each dimension of $\mathbf{p}^{\mathcal{I}}_{i,j}$ denotes the probability of a tag between $t_i$ and $t_j$. 
The sigmoid function is used to transfer the projected value into a probability, in this case, the cross-entropy loss can be used as the loss function which has been proved suitable for multi-label classification task:
\begin{align}
\mathcal{L}^{\mathcal{I}}=-&\sum_{i=1}^n\sum_{j=s^{\mathcal{I}}}^n\sum_{k=1}^{K^{\mathcal{I}}} (\mathbf{y}_{i,j}^{\mathcal{I}}[k]{\rm log}(\mathbf{p}^{\mathcal{I}}_{i,j}[k])\\&+(1-\mathbf{y}_{i,j}^{\mathcal{I}}[k]){\rm log}(1-\mathbf{p}^{\mathcal{I}}_{i,j}[k])), \nonumber
%    \mathbf{o}^s_{i:j} = \mathbf{W}^\alpha_o \cdot \mathbf{h}^s_{i:j} + \mathbf{b}^\alpha_o, 
\end{align}
where $K^{\mathcal{I}}$ is the number of pre-defined tags in $\mathcal{I}$, $\mathbf{p}^{\mathcal{I}}_{i,j}[k] \in [0,1]$ is the predicted probability of $(t_i,t_j)$ along the $k$-th tag, and $\mathbf{y}_{i,j}^{\mathcal{I}}[k]\in \{0,1\}$ is the corresponding ground truth.
$s^{\mathcal{I}}$ equals to $1$ if $\mathcal{I}=\text{e}$ or $i$ if $\mathcal{I}=\text{s}$. 
Then, the losses from segment extraction and edge prediction are aggregated to form the training objective $\mathcal{J}(\theta)$:
\begin{equation}
\mathcal{J}(\theta)=\mathcal{L}^{\text{s}}+\mathcal{L}^{\text{e}}.
\end{equation}

At inference, the probability vector $\mathbf{p}^{\mathcal{I}}_{i,j}$ needs thresholding to be converted to tags.
We enumerate several values in the range $(0, 1)$ and pick the one that maximizes the evaluation metrics on the validation (dev) set as the threshold.

%The RE classifier outputs the probability P (r|es, eo) within the range [0, 1], which needs thresholding to be converted to relation labels.

%enumerating several values in the range (0, 1) and picking the one that maximizes the evaluation metrics

%Inspired by \newcite{sun2020circle} and \newcite{multilabel_crossentropy}, we define the training loss as below to handle this multi-label classification problem:
%\begin{align}
%    \label{equ:loss_func}
%    L_{pos} &= \log (1 + \sum_{k \in Y_{i,j}} \exp^{-{o}^*_{i,j,k}}) \\
%    L_{neg} &= \log (1 + \sum_{k \notin Y_{i,j}} \exp^{{o}^*_{i,j,k}}) \\
%    L &= L_{pos} + L_{neg}
%\end{align}
%
%Here, ${o}^*_{i,j,k}$ is the k-th element of the output vector ${o}^*_{i,j}$. $Y_{i,j}$ denotes the gold tag set of the token pair i-j. Since $\log (\sum_{x \in X} \exp^x)$ is the smooth approximation to $\max (X)$, we can make positive tags greater than 0 and negative tags less than 0 by minimizing the loss function.
\section{Evaluation}
% In this section, after introducing the datasets and baseline models, we present our experimental results and detailed analysis.

% report the performance of Mac on the whole dataset. To further study the capability in extracting discontinuous entity mentions and the robustness to overlapping structure, segment length, and interval length, we also report its performance on corresponding test data subsets. 

\subsection{Datasets}
Following previous work~\cite{dai-etal-2020-effective}, we conduct experiments on three benchmark datasets from the biomedical domain: (1) CADEC~\cite{karimi2015cadec} is sourced from AskaPatient: an online forum where patients can discuss their experiences with medications. 
We use the dataset pre-processed by ~\citeauthor{dai-etal-2020-effective}\shortcite{dai-etal-2020-effective} which selected Adverse Drug Event (ADE) annotations from the original dataset because only the ADEs involve discontinuous annotations.
(2) ShARe 13~\cite{pradhan2013task} and (3) ShARe 14~\cite{mowery2014task} focus on the identification of disorder mentions in clinical notes, including discharge summaries, electrocardiogram, echocardiogram, and radiology reports.
Around 10\% of mentions in these three data sets are discontinuous. The descriptive statistics of the datasets are reported in Table~\ref{table:statistics_main}.

\subsection{Implementation Details}\label{sec:implementation details}
We implement our model upon the in-field BERT base model: Yelp Bert~\cite{dai2020cost} for CADEC, and Clinical BERT~\cite{alsentzer2019publicly} for ShARe 13 and 14. 
The network parameters are optimized by Adam~\cite{kingma2014adam} with a learning rate of 1e-5.
% using PyTorch~\cite{paszke2019pytorch} and the network parameters are optimized by Adam~\cite{kingma2014adam} with a learning rate of 1e-5.
% We try 
% Our mode
% two kinds of encoders in this paper.
% One is the general-domain BERT base model~\cite{devlin2018bert}, and another is the in-field BERT: Yelp Bert~\cite{dai2020cost} for CADEC, and Clinical BERT~\cite{alsentzer2019publicly} for ShARe 13 and 14. 
% The resulting two models are named as Mac$_{B}$ and Mac$_{B^*}$, respectively.
% For learning rate schedule, we choose the Cosine Annealing Warm Restarts~\cite{loshchilov2016sgdr}. 
% We conduct Cosine Annealing Warm Restarts learning rate schedule~\cite{loshchilov2016sgdr}. 
% The max length of input sentence is set to 64/100 for training/evaluation respectively.
The batch size is fixed to 12.
The threshold for converting probability to tag is set as 0.5.
% The batch size is set as 12. 
All the hyper-parameters are tuned on the dev set.
We run our experiments on a NVIDIA Tesla V100 GPU for at most 300 epochs, and choose the model with the best performance on the dev set to output results on the test set. 
we report the test score of the run with the median dev score among 5 randomly initialized runs.
% We repeat the training procedure five times with different random seeds, and report the median result of them. 

\begin{table}[t]
\small
\begin{center}
    \setlength{\tabcolsep}{0.45mm}{\begin{tabular}{lcccccccccc}
    \toprule
        \multirow{2}{*}{} & 
        \multicolumn{3}{c}{CADEC} & 
        \multicolumn{3}{c}{ShARe 13} & 
        \multicolumn{3}{c}{ShARe 14} &
        \\
        & train & dev & test
        & train & dev & test
        & train & dev & test
        \\
    \midrule
        S
        & 5,340 & 1,097 & 1,160 
        & 8,508 & 1,250 & 9,009 
        & 17,407 & 1,361 & 15,850
        \\
        M 
        & 4,430 & 898 & 990 
        & 5,146 & 669 & 5,333 
        & 10,354 & 771 & 7,922
        \\
        D 
        & 491 & 94 & 94 
        & 581 & 71 & 436 
        & 1,004 & 80 & 566
        \\
        P
        & 11.1 & 10.5 & 9.5 
        & 11.3 & 10.6 & 8.2 
        & 9.7 & 10.4 & 7.1
        \\
    \bottomrule
    \end{tabular}}
{
    \caption{Statistics of datasets. S, M, and D respectively represent the number of sentences, total mentions, and discontinuous mentions. P denotes the percentage of discontinuous mentions in total mentions.}
    \label{table:statistics_main}
} 
\end{center}
\end{table}

\begin{table}[t]
\small
\begin{center}
    \setlength{\tabcolsep}{0.7mm}{
    \begin{tabular}{lccccccccc}
    \toprule
        \multirow{2}{*}{Model} & 
        \multicolumn{3}{c}{CADEC} &  \multicolumn{3}{c}{ShARe 13} & \multicolumn{3}{c}{ShARe 14}
        \\
        % & P & R & F 
        % & P & R & F 
        % & P & R & F 
        & Prec. & Rec. & F1 
        & Prec. & Rec. & F1 
        & Prec. & Rec. & F1 
        \\
    \midrule
        BIOE
        & 68.7 & 66.1 & 67.4 & 77.0 & 72.9 & 74.9 & 74.9  & 78.5 & 76.6 \\
        Graph
        & 72.1 & 48.4 & 58.0 & 83.9 & 60.4 & 70.3 & \textbf{79.1} & 70.7 & 74.7 \\
        Comb$_B$ % reproduced
        & 69.8 & 68.7 & 69.2 & 80.1 & 73.9 & 76.9 & 76.5 & 82.3 & 79.3  \\
        Trans$_E$
        & 68.9 & 69.0 & 69.0 & 80.5 & 75.0 & 77.7 & 78.1 & 81.2 & 79.6  \\
    % \midrule
        % Trans$_E^\dagger$ % reproduced
        % & 69.8 & 68.7 & 69.2 & 80.1 & 73.9 & 76.9 & 76.5 & 82.3 & 79.3  \\
        % Trans$_B$
        % & 69.2 & 66.0 & 67.6 & 74.5 & 72.9 & 73.7 & 75.9 & 77.1 & 76.5  \\
        Trans$_{B}$
        & 68.8 & 67.3 & 68.0 & 77.3 & 72.9 & 75.0 & 76.0 & 78.6 & 77.3  \\
    \midrule
        % Mac$_B$ 
        % & 70.1 & 71.0 & 70.5 & 81.5 & 75.2 & 78.2 & \textbf{79.1} & 80.4 & 79.7 \\
        Mac 
        & \textbf{70.5} & \textbf{72.5} & \textbf{71.5} & \textbf{84.3} & \textbf{78.2} & \textbf{81.2} & 78.2 & \textbf{84.7} & \textbf{81.3} \\
    \bottomrule
    \end{tabular}}
{
    \caption{ Main results on three benchmark datasets. Bold marks highest number among all models.
    }
    \label{table:main_res}
} 
\end{center}
\end{table}

\subsection{Comparison Models}
For comparison, we employ the following models as baselines: (1) BIOE~\cite{metke2016concept} expands the BIO tagging scheme with additional tags to represent discontinuous entity; (2) Graph~\cite{muis2016learning} uses hyper-graphs to organize entity spans and their combinations; (3) Comb~\cite{wang2019combining} first detects entity spans, then deploys a classifier to merge them.
For fair comparison, we re-implement Comb based on the in-fild BERT backbone called Comb$_B$; (4) Trans$_E$~\cite{dai-etal-2020-effective} is the current best discontinuous NER method, which generates a sequence of actions with the aid of buffer and stack structure to detect entity;
Note that the original Trans$_E$ model is based on ELMo.
For fair comparison with our model, we also implement the in-field BERT-based Trans models, namely Trans$_B$.

\begin{table*}[t]
\small
\begin{center}
    \setlength{\tabcolsep}{1.5mm}{
    \begin{tabular}{lccccccccccccc}
    \toprule
        \multirow{2}{*}{Model} 
        & \multicolumn{3}{c}{CADEC} 
        &  \multicolumn{3}{c}{ShARe 13} 
        & \multicolumn{3}{c}{ShARe 14}
        \\
         & Prec. & Rec. & F1 & Prec. & Rec. & F1 & Prec. & Rec. & F1 \\
    \midrule
        BIOE 
        & 68.3/~5.8 & 52.0/~1.0 & 57.3/~1.8 & 51.8/~39.7 & 39.5/~12.3 & 44.8/~18.8 & 37.5/~8.8  & 38.4/~4.5 & 37.9/~6.0 \\
        Graph 
        & 69.5/~\textbf{60.8} & 43.2/~14.8 & 53.3/~23.9 & \textbf{82.3}/~78.4 & 47.4/~36.6 & 60.2/~50.0 & 60.0/~42.7 & 52.8/~39.5 & 56.2/~41.1 \\
        Comb$_{B}$ 
        & 63.9/~44.0 & 57.8/~23.4 & 60.7/~30.6 & 59.7/~65.5 & 49.8/~29.6 & 54.3/~40.8 & 52.9/~51.2 & 52.8/~35.0 & 52.9/~41.6 \\
        Trans$_E$ 
        & 66.5/~41.2 & 64.3/~35.1 & 65.4/~37.9 & 70.5/~\textbf{78.5} & 56.8/~39.4 & 62.9/~52.5 & 61.9/~\textbf{56.1} & 64.5/~43.8 & 63.1/~49.2  \\
    % \midrule 
    %     Trans$_E^\dagger$ % reproduced
    %     & 65.6/~38.9 & 59.6/~29.8 & 62.5/~33.7 & 65.6/~68.7 & 53.5/~39.7 & 58.9/~50.3 & 58.1/~51.4 & 60.0/~44.5 & 59.0/~47.7  \\
        % Trans$_B$
        % & 63.9/~44.0 & 57.8/~23.4 & 60.7/~30.6 & 59.7/~65.5 & 49.8/~29.6 & 54.3/~40.8 & 52.9/~51.2 & 52.8/~35.0 & 52.9/~41.6  \\
        Trans$_{B}$
        & 69.1/~39.5 & 64.4/~34.0 & 66.7/~36.6 & 68.2/~65.9 & 55.4/~39.0 & 61.1/~49.0 & 55.5/~52.0 & 55.6/~37.8 & 55.6/~43.8  \\
    \midrule
        % Mac$_B$ 
        % & 68.8/~53.7 & 64.0/~31.0 & 66.3/~39.2 & 77.4/~70.0 & 58.8/~46.1 & 66.9/~55.6 & \textbf{71.0}/~55.2 & 64.3/~50.9 & 67.5/~53.0 \\
        Mac 
        & \textbf{74.7}/~52.9 & \textbf{65.5}/~\textbf{38.3} & \textbf{69.8}/~\textbf{44.4} & 77.9/~66.1 & 60.5/~\textbf{48.4} & \textbf{68.1}/~\textbf{55.9} & 69.3/~51.0 & \textbf{70.2}/~\textbf{57.6} & \textbf{69.7}/~\textbf{54.1} \\
    \bottomrule
    \end{tabular}}
{
    \caption{Results on discontinuous entity mentions. In the Table, two scores are reported and separated by a slash (``/''). The former is the score on sentences with at least one discontinuous entity mention. The latter is the score only considering discontinuous entity mentions.
    }
    \label{table:disc_res}
} 
\end{center}
\end{table*}

\subsection{Main results}
% \subsubsection{Main Results}
% overall 

Table~\ref{table:main_res} reports the results of our model against other baseline methods. 
We have the following observations.
(1) Our method, Mac, significantly outperforms all other methods and achieves the SOTA F1 score on all three datasets.
(2) BERT-based Trans model achieves poorer results than its ELMo-based counterpart, which is in line with the claim in the original paper.
(3) Over the SOTA method Trans$_{E}$, Mac achieves substantial improvements of 2.6\% in F1 score on three datasets averagely.
Moreover, the Wilcoxon’s test shows that a significant difference ($p < 0.05$) exists between our model and Trans$_{E}$.
We consider that it is because Trans$_{E}$ is inherently a multi-stage method as it introduces several dependent actions, thus suffering from the exposure bias problem.
While for our Mac method, it elegantly decomposes the discontinuous NER task into two independent subtasks and learns them together with a joint model, realizing the consistency of training and inference.
(4) Comb$_{B}$ can be approximately seen as the pipeline version of our method, their performance gap again confirms the effectiveness  of our one-stage learning framework.

% The first three lines of results are quoted directly from the original papers. Since our Mac is based on BERT, for fair comparisons, we re-implement the transition-based model\cite{dai-etal-2020-effective} with the BERTs used in Mac and report corresponding results. The results are in line with the claim in their paper, the transition-based models with BERT achieve poorer performance than the one with ELMo. As shown Table~\ref{table:main_res}, our Mac outperforms the three baseline models on all metrics.

% disc
As shown in Table~\ref{table:statistics_main}, only around 10\% mentions are discontinuous in all three datasets, which is far less than the continuous entity mentions. 
To evaluate the effectiveness of our proposed model on recognizing discontinuous mentions, following \newcite{muis2016learning}, we report the results on sentences that include at least one discontinuous mention. 
We also report the evaluation results when only discontinuous mentions are considered.
The scores in these two settings are separated by a slash in Table~\ref{table:disc_res}.
% Further, we also report the results that only discontinuous mentions are considered. 
% In Table~\ref{table:disc_res}, the two scores separated by a slash (``/'') are the score on sentences with at least one discontinuous mention, and the score only considering discontinuous mentions. 
% BIO EXT
Comparing Table~\ref{table:main_res} and \ref{table:disc_res}, we can see that the BIOE model performs better than the Graph when testing on the full dataset but far worse on discontinuous mentions. 
% This phenomenon suggests that achieving higher scores on benchmarks does not mean better performance on discontinuous NER task.
Consistently, our model again defeat the baseline models in terms of F1 score. Even though some models outperform Mac on precision or recall, they greatly sacrifice another score, which results in lower F1 score than Mac.

% discontinuous entity

% further analysis
% overlapping patterns
\subsection{Model Ablation Study}

\begin{table}[t]
\small
\begin{center}
    \setlength{\tabcolsep}{1.5mm}{\begin{tabular}{lcccc}
    \toprule
        Model
        & F1
        & Dis F1
        & Dis F1$^\star$ \\
    \midrule
        Mac & \textbf{78.7} & \textbf{56.4} & \textbf{46.6} \\
        ~~-- Tag B and S & 78.2 & 55.8 & 46.1 \\
        ~~-- Segment length embedding & 78.1 & 55.7 & 46.2 \\
        ~~-- CLN mechanism & 76.8 & 52.7 & 44.4 \\
        ~~-- Segment inner representation & 72.9 & 55.6 & 46.3 \\
         
    \bottomrule
    \end{tabular}}
{
    \caption{An ablation study on the ShARe 13 dev set. F1, Dis F1, and Dis F1$^\star$ respectively denote the overall F1 score, F1 score on sentences with at least one discontinuous mention, and on discontinuous mentions. }
    \label{table:ablation}
} 
\end{center}
\end{table}

To verify the effectiveness of each component, we ablate one component at a time to understand its impact on the performance. Concretely, we investigated the tagging scheme of segments, the segment length embedding, the CLN mechanism (by replacing it with the vector concatenation), and the segment inner token representation.

From these ablations shown in Table~\ref{table:ablation}, we find that: (1) When we take B, I and S tags in segment extraction as one class, the score slightly drops by 0.5\%, which indicates the segments in different positions of entities may have different semantic features, so distinguishing them can reduce the confusion in the process of model recognition; (2) When we remove the segment length embedding (Formula~\ref{equ:hsk_seg_len_emb}), the overall F1 score drops by 0.6\%, showing that it is necessary to let segment extractor aware of the token pair distance information to filter out impossible segments by implicit distance constraint; (3) Compared with concatenating, it is a better choice to use CLN (Formula~\ref{equ:hsk_seg_cln} and \ref{equ:int_rel_cln}) to fuse the features of two tokens, which brings 1.9\% improvement; (4) Removing segment inner features (Formula~\ref{equ:hsk_seg_lstm}) results in a remarkable drop on the overall F1 score while little drop on the scores of discontinuous mentions, which suggests that the information of inner tokens is essential to recognize continuous entity mentions. 
Overall, we can conclude that the improvement of grid encoder brings significant performance gains.

\begin{figure}[t]
    \centering
    \includegraphics[width=0.95\columnwidth]{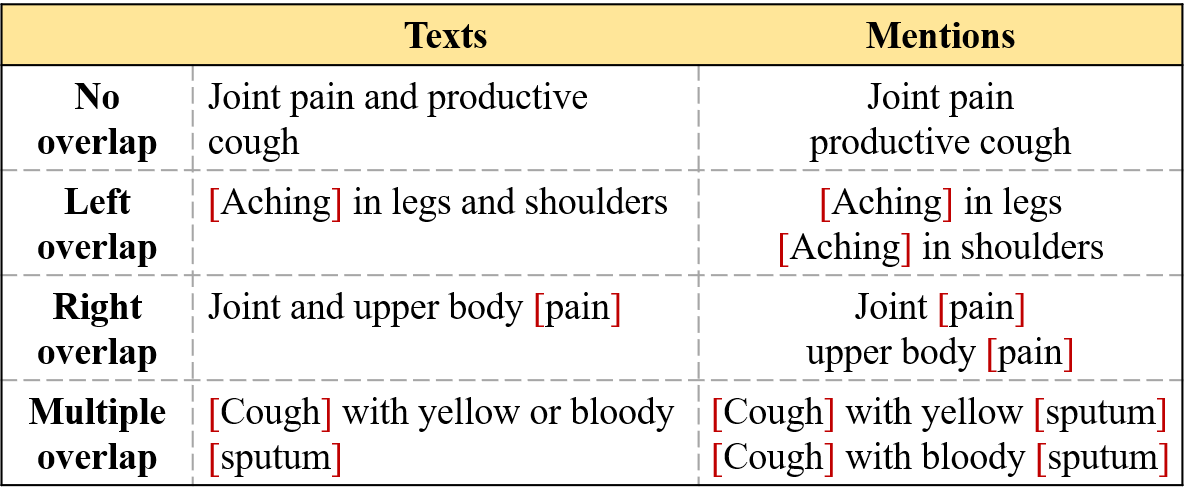}
    \caption{Examples of the overlapping patterns}
    \label{fig:examples_overlap_patterns}
    \vspace{-0.1in}
\end{figure}

\begin{table}[t]
\small
\begin{center}
    \setlength{\tabcolsep}{1.mm}{\begin{tabular}{lcccccccccc}
    \toprule
        \multirow{2}{*}{Pattern} & 
        \multicolumn{3}{c}{CADEC} & 
        \multicolumn{3}{c}{ShARe 13} & 
        \multicolumn{3}{c}{ShARe 14} &
        \\
        & train & dev & test
        & train & dev & test
        & train & dev & test
        \\
    \midrule
        No 
        & 57 & 9 & 16 
        & 348 & 41 & 193 
        & 535 & 39 & 246
        \\
        Left 
        & 270 & 54 & 41 
        & 167 & 11 & 200 
        & 352 & 30 & 238
        \\
        Right 
        & 113 & 16 & 23 
        & 48 & 19 & 35 
        & 97 & 5 & 67
        \\
        Multi. 
        & 51 & 15 & 14 
        & 18 & 0 & 8 
        & 20 & 6 & 15
        \\
    \bottomrule
    \end{tabular}}
{
    \caption{Statistics of overlapping patterns.}
    \label{table:statistics_overlap}
} 
\end{center}
\end{table}

\begin{figure*}[t]
\centering
\subfigure[CADEC]{
    \begin{minipage}[t]{0.33\linewidth}
    \centering
    % \label{fig:fact_num}
    \includegraphics[scale=0.3]{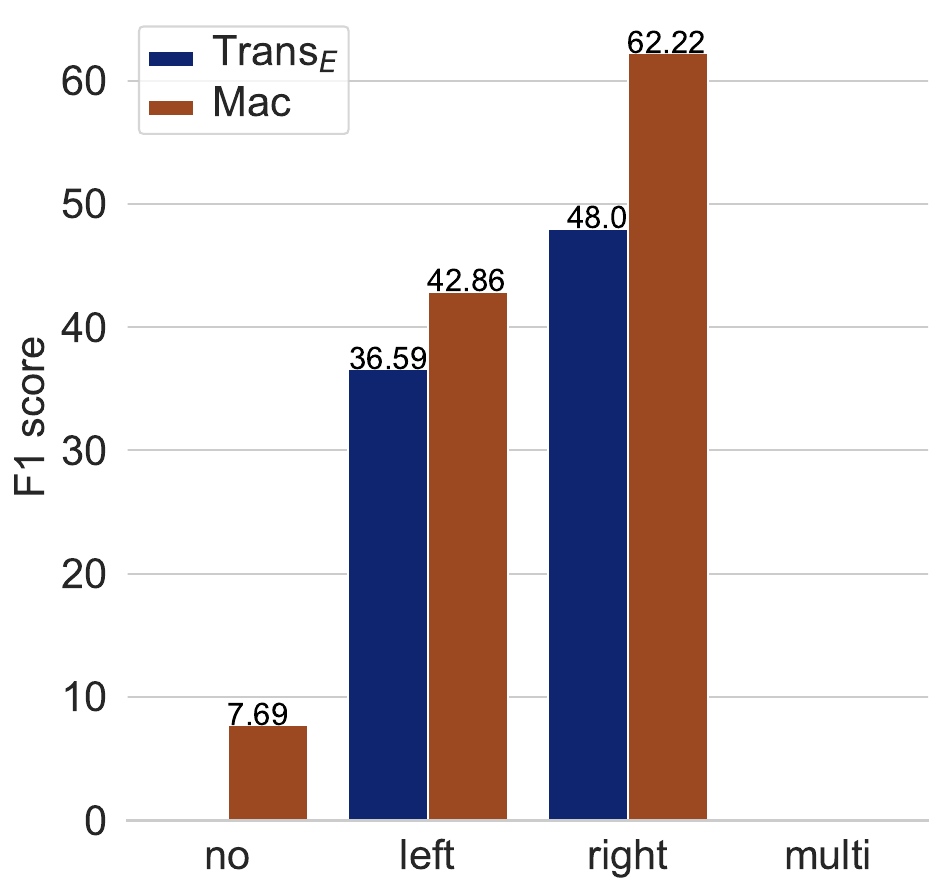}
    \end{minipage}%
}%
\subfigure[ShARe 13]{
    \begin{minipage}[t]{0.33\linewidth}
    % \label{fig:discontinuous_type}
    \centering
    \includegraphics[scale=0.3]{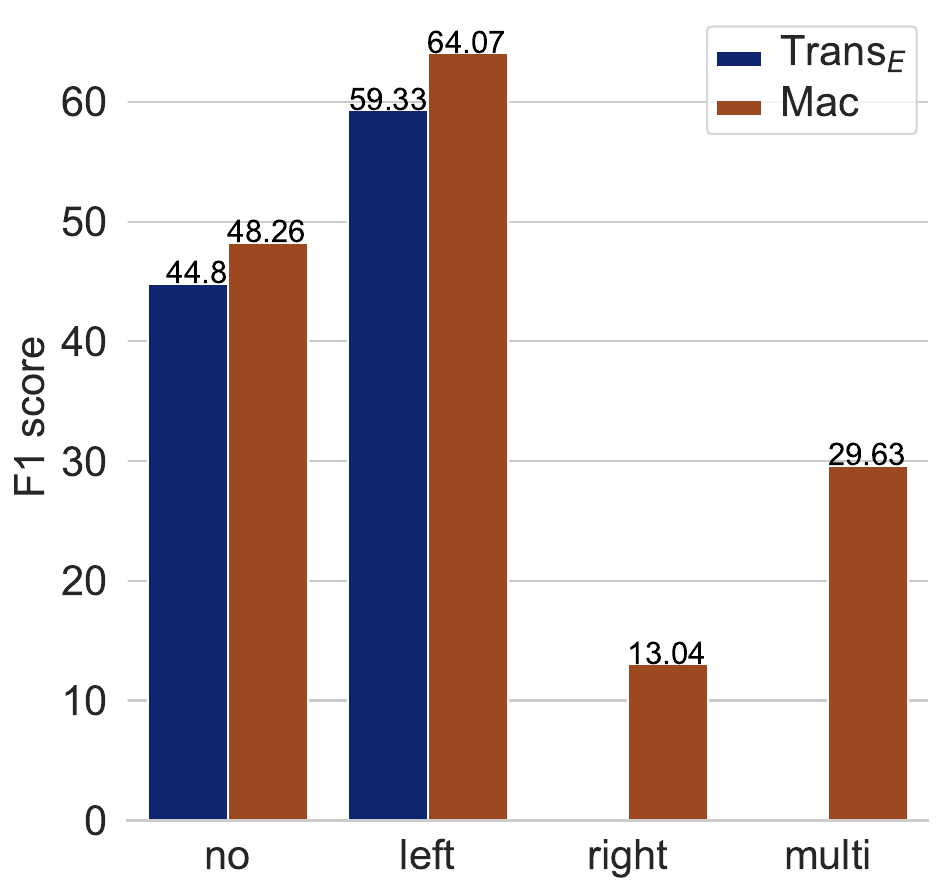}
    \end{minipage}%
}%
\subfigure[ShARe 14]{
    \begin{minipage}[t]{0.33\linewidth}
    % \label{fig:unseen}
    \centering
    \includegraphics[scale=0.3]{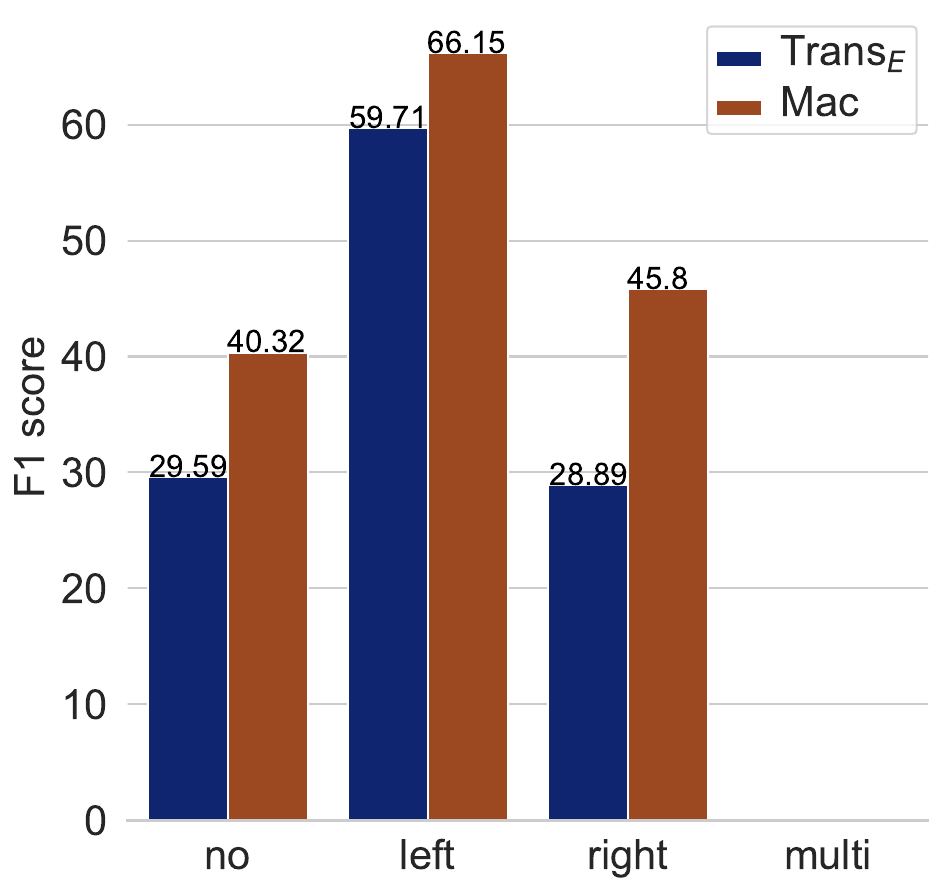}
    \end{minipage}
}%
\centering
\caption{ Performance on different overlapping patterns.}
\label{fig:res_overlap_patterns}
\end{figure*}

% footnote{Limited by the space, we analyze the performance with regard to different interval and span length in Appendix.}

\subsection{Performance Analysis}

\subsubsection{Impact of Overlapping Structure}

% \begin{table*}[t]
% \small
% \begin{center}
%     \setlength{\tabcolsep}{1.5mm}{
%     \begin{tabular}{lccccccccccccc}
%     \toprule
%         \multirow{2}{*}{Model} 
%         & \multicolumn{3}{c}{CADEC} 
%         &  \multicolumn{3}{c}{ShARe 13} 
%         & \multicolumn{3}{c}{ShARe 14}
%         \\
%          & Prec. & Rec. & F1 & Prec. & Rec. & F1 & Prec. & Rec. & F1 \\
%     \midrule
%         No 
%         & 0/~\textbf{10.0} & 0/~\textbf{6.3} & 0/~\textbf{7.7} & \textbf{57.3}/~55.0 & 36.8/~\textbf{43.0} & 44.8/~\textbf{48.3} & 39.7/~\textbf{40.0} & 23.5/~\textbf{40.6} & 29.6/~\textbf{40.3} \\
        
%         Left 
%         & 36.6/~\textbf{51.7} & 36.6/~36.6 & 36.6/~\textbf{42.9} & 76.3/~\textbf{79.9} & 48.5/~\textbf{53.5} & 59.3/~\textbf{64.1} & 52.2/~\textbf{61.0} & 69.7/~\textbf{72.3} & 59.7/~\textbf{66.2} \\
        
%         Right
%         & 44.4/~\textbf{63.6} & 52.1/~\textbf{60.9} & 48.0/~\textbf{62.2} & 0/~\textbf{27.3} & 0/~\textbf{8.6} & 0/~\textbf{13.0} & \textbf{56.5}/~46.9 & 19.4/~\textbf{44.8} & 28.9/~\textbf{45.8} \\
        
%         Multiple
%         & 0/~0 & 0/~0 & 0/~0 & 0/~\textbf{21.1} & 0/~\textbf{50.0} & 0/~\textbf{29.6} & 0/~0 & 0/~0 & 0/~0  \\
        
%     \bottomrule
%     \end{tabular}}
% {
%     \caption{Performance on different overlapping patterns. In the Table, the scores of $Trans_E$ and $Mac$ are reported and separated by a slash (``/'').
%     }
%     \label{table:res_overlap_patterns}
% } 
% \end{center}
% \end{table*}

As discussed in the introduction, overlap is very common in discontinuous entity mentions.
To evaluate the capability of our model on extraction overlapping structures, as suggested in ~\cite{dai-etal-2020-effective}, we divide the test set into four categories: (1) no overlap; (2) left overlap; (3) right overlap; and (4) multiple overlap. Figure~\ref{fig:examples_overlap_patterns} gives examples for each overlapping pattern. 
As illustrated in Figure~\ref{fig:res_overlap_patterns}, Mac outperforms Trans$_{E}$ on all the overlapping patterns. 
Trans$_{E}$ gets zero scores on some patterns. 
It might result from insufficient training since these overlapping patterns have relatively fewer samples in the training sets (see Table~\ref{table:statistics_overlap}), while the sequential action structure of transition-based model is a bit data hungry. 
By contrast, Mac is more resilient to overlapping patterns,
we attribute the performance gains to two design choices:
(1) the grid tagging scheme has strong power in accurately identifying overlapping segments and assembling them into a segment graph;
(2) Based on the graph, the maximal clique discovery algorithm can effectively recover all the candidate overlapping entity mentions.

\subsubsection{Impact of Interval and Span Length}
% Performance on Different Interval and Span Lengths

% interval length
\begin{table}[t]
\small
\begin{center}
    \setlength{\tabcolsep}{1.mm}{\begin{tabular}{lcccccccccc}
    \toprule
        \multirow{2}{*}{Length} & 
        \multicolumn{3}{c}{CADEC} & 
        \multicolumn{3}{c}{ShARe 13} & 
        \multicolumn{3}{c}{ShARe 14} &
        \\
        & train & dev & test
        & train & dev & test
        & train & dev & test
        \\
    \midrule
        = 1 
        & 36 & 8 & 8 
        & 96 & 15 & 125 
        & 227 & 10 & 107
        \\
        = 2
        & 217 & 42 & 54 
        & 215 & 26 & 118 
        & 322 & 33 & 146
        \\
        = 3
        & 56 & 14 & 12 
        & 102 & 12 & 91 
        & 184 & 20 & 120
        \\
        = 4 
        & 68 & 14 & 8 
        & 46 & 3 & 16 
        & 61 & 3 & 43
        \\
        = 5 
        & 36 & 4 & 4 
        & 48 & 4 & 46 
        & 92 & 6 & 61
        \\
        = 6
        & 30 & 3 & 3 
        & 25 & 3 & 12 
        & 38 & 2 & 31
        \\
        $\geq$ 7
        & 48 & 9 & 5 
        & 49 & 8 & 28 
        & 80 & 6 & 58
        \\
    \bottomrule
    \end{tabular}}
{
    \caption{Statistics of interval length.}
    \label{table:statistics_interval_len}
} 
\end{center}
\end{table}

% span length
\begin{table}[t]
\small
\begin{center}
    \setlength{\tabcolsep}{1.mm}{\begin{tabular}{lcccccccccc}
    \toprule
        \multirow{2}{*}{Length} & 
        \multicolumn{3}{c}{CADEC} & 
        \multicolumn{3}{c}{ShARe 13} & 
        \multicolumn{3}{c}{ShARe 14} &
        \\
        & train & dev & test
        & train & dev & test
        & train & dev & test
        \\
    \midrule
        = 3
        & 10 & 3 & 4 
        & 30 & 7 & 93 
        & 124 & 6 & 74
        \\
        = 4
        & 95 & 23 & 24 
        & 108 & 25 & 71 
        & 190 & 15 & 113
        \\
        = 5
        & 67 & 13 & 15 
        & 157 & 17 & 115 
        & 259 & 27 & 140
        \\
        = 6 
        & 91 & 13 & 16 
        & 125 & 3 & 51 
        & 165 & 12 & 65
        \\
        = 7 
        & 57 & 15 & 9 
        & 65 & 5 & 61 
        & 120 & 10 & 76
        \\
        = 8
        & 53 & 9 & 10 
        & 27 & 4 & 14 
        & 42 & 3 & 33
        \\
        $\geq$ 9
        & 118 & 18 & 16 
        & 69 & 10 & 31 
        & 104 & 7 & 65
        \\
    \bottomrule
    \end{tabular}}
{
    \caption{Statistics of span length.}
    \label{table:statistics_span_len}
} 
\end{center}
\end{table}

\begin{figure}[t]
\centering
\caption{Performance on different interval length.}
\label{fig:inter_len}
\subfigure[CADEC]{
    \begin{minipage}[t]{0.45\linewidth}
    \centering
    \includegraphics[scale=0.25]{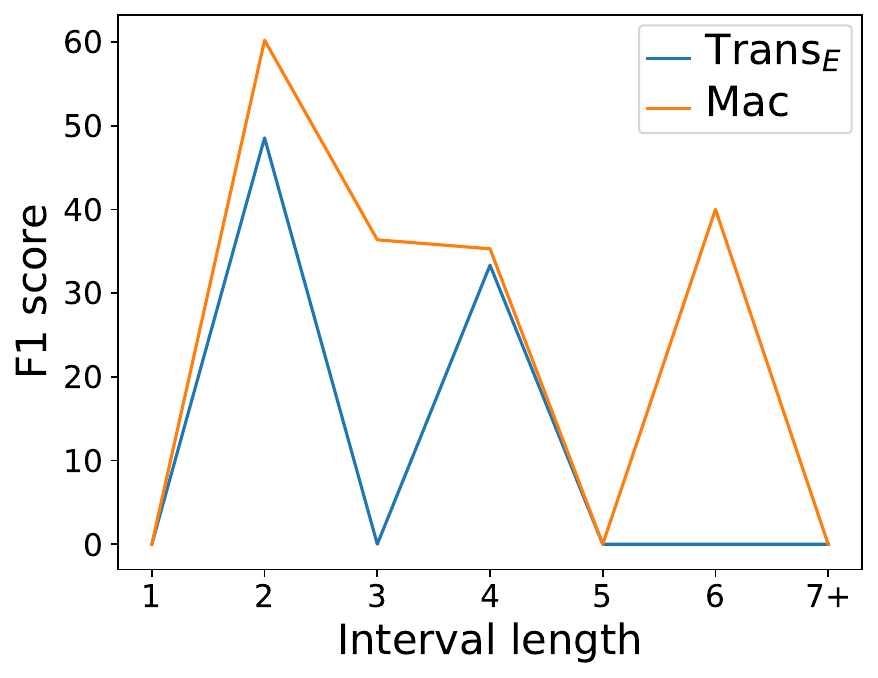}
    \end{minipage}
}
\subfigure[CADEC]{
    \begin{minipage}[t]{0.45\linewidth}
    \centering
    \includegraphics[scale=0.25]{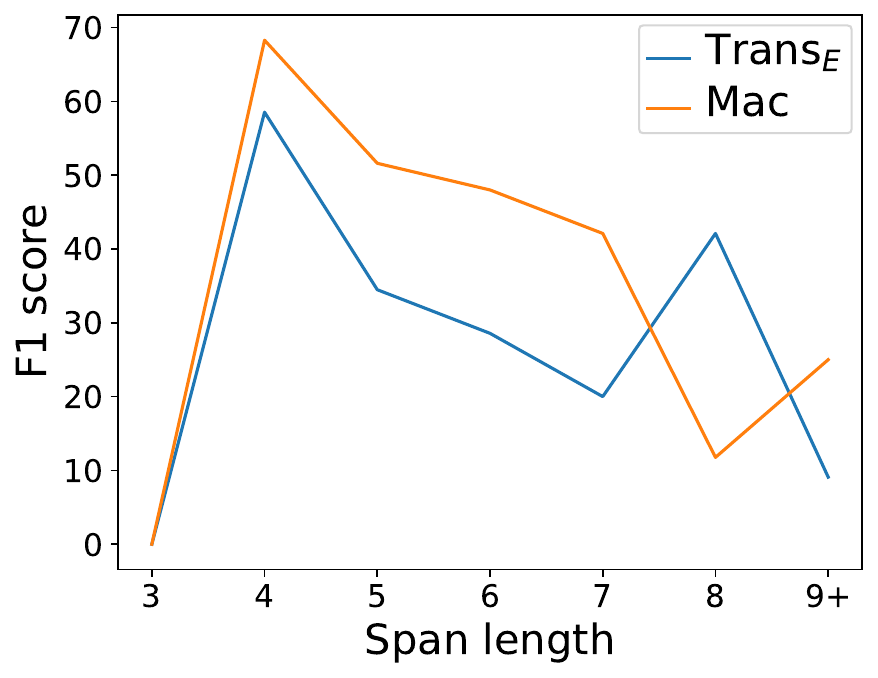}
    \end{minipage}
}

\subfigure[ShARe 13]{
    \begin{minipage}[t]{0.45\linewidth}
    \centering
    \includegraphics[scale=0.25]{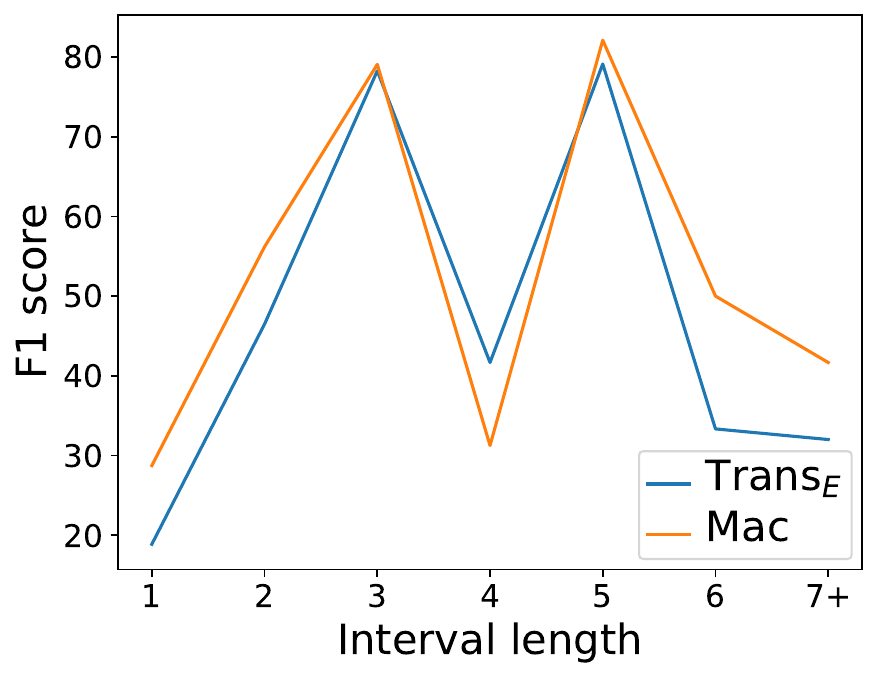}
    \end{minipage}
}
\subfigure[ShARe 13]{
    \begin{minipage}[t]{0.45\linewidth}
    \centering
    \includegraphics[scale=0.25]{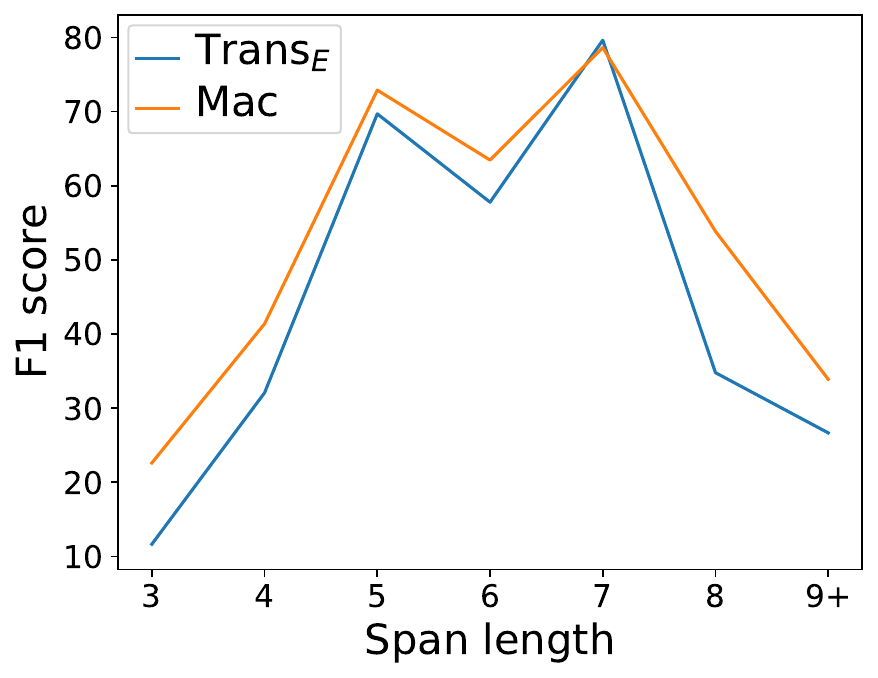}
    \end{minipage}
}

\subfigure[ShARe 14]{
    \begin{minipage}[t]{0.45\linewidth}
    \centering
    \includegraphics[scale=0.25]{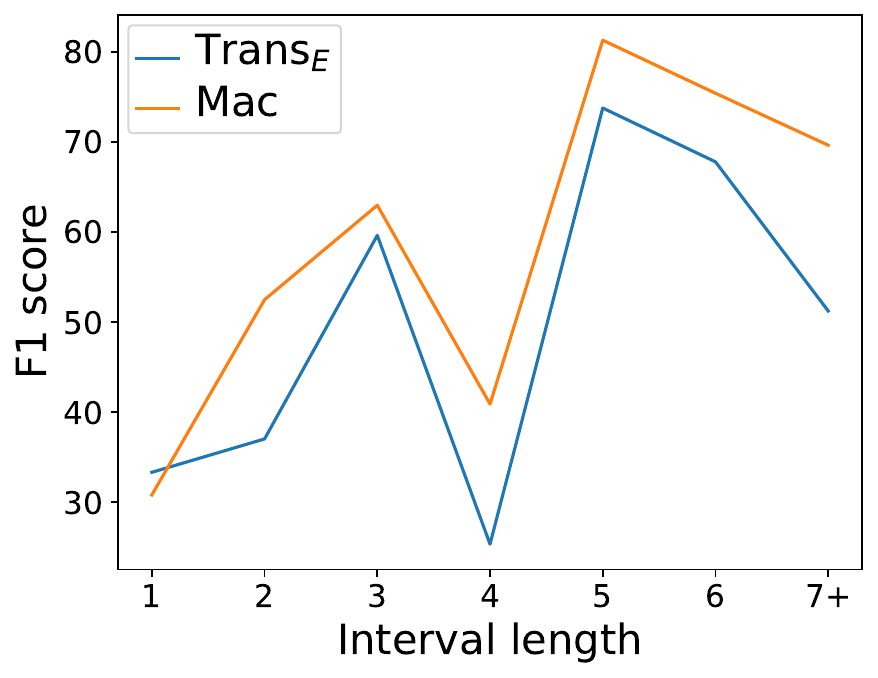}
    \end{minipage}
}
\subfigure[ShARe 14]{
    \begin{minipage}[t]{0.45\linewidth}
    \centering
    \includegraphics[scale=0.25]{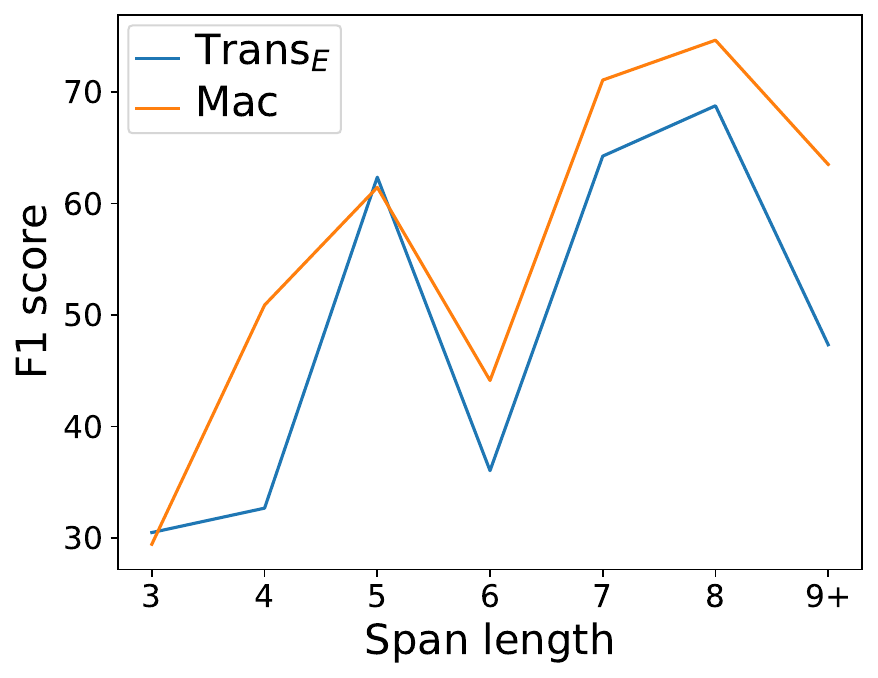}
    \end{minipage}
}
\centering
\end{figure}

Intervals between segments usually make the total length of a discontinuous mention longer than continuous one. Considering the involved segments, the whole span is even longer. That is, different words of a discontinuous mention may be distant to each other, which makes discontinuous NER harder than the conventional NER task. To further evaluate the robustness of Mac in different settings, we analyze the results of test sets on different interval and span lengths. The interval length refers to the number of words between discontinuous segments. The span length refers to the number of words of the whole span. For example, for the entity mention ``\textit{Sever shoulder pain}'' in ``\textit{\textbf{Sever} joint, \textbf{shoulder} and upper body \textbf{pain}.}'', the interval length is 5, and the span length is 8. Such phenomenon requires models to have the ability of capturing the semantic dependency between distant segments. 

For the convenience of analysis, we report all datasets' distribution on interval and span length in Table~\ref{table:statistics_interval_len} and \ref{table:statistics_span_len}, respectively. And Figure~\ref{fig:inter_len} shows the F1 scores of Trans$_E$ and Mac on different interval and span lengths. 
As we can see, Mac outperforms Trans$_E$ in most setting. Even though Mac is defeated in some cases, the sample number in those cases is too small to disprove the superiority of Mac. For example, on CADEC, Trans$_E$ outperforms Mac when span length is 8, but the sample number in the test set is only 10.

We figure out an interesting phenomenon: Both Mac and Trans$_E$ show poor performance when interval length is 1 and span length is 3, even though the corresponding training samples are sufficient enough (see length = 1 in Table~\ref{table:statistics_interval_len} and length = 3 in Table~\ref{table:statistics_span_len}\footnote{For discontinuous mentions, when span length is 3, the interval length can only be 1.}). 
% For example, ShARe 14 has over 200 training samples, of which the interval length is 1, but both models perform much worse than when interval length is 3, which has less training samples. 
This might result from two folds: (1) Even though the training samples are sufficient, their features and context are  different from the ones in the test set; (2) discontinuous mentions with interval length equal to 1 are harder cases than the others, since only one word to separate the segments makes these discontinuous mentions very similar to the continuous ones, which confuse the model to treat them as a continuous mention. We leave this problem to our future work.

%ablation study

% speed
\subsubsection{Analysis on Running Speed}

\begin{table}[t]
\small
\begin{center}
    \setlength{\tabcolsep}{1.5mm}{\begin{tabular}{lcccc}
    \toprule
        Model
        & CADEC
        & ShARe 13
        & ShARe 14\\
    \midrule
        Trans$_{B}$ & 29.1 Sen/s & 33.4 Sen/s & 33.9 Sen/s \\
        Trans$_{E}$ & 36.3 Sen/s & 40.6 Sen/s & 40.3 Sen/s \\
        Mac & 193.3 Sen/s & 200.2 Sen/s & 198.1 Sen/s \\
    \bottomrule
    \end{tabular}}
{
    \caption{Comparison on running speed. Sen/s refers to the number of sentences can be processed per second. }
    \label{table:computational_eff}
} 
\end{center}
\end{table}

Table~\ref{table:computational_eff} shows the comparison of computational efficiency between the SOTA model Trans$_{E}$, Trans$_{B}$, and our proposed Mac. 
% In this experiment, we use the officially open-sourced code of Trans and replace its encoder with in-field BERT (the same as Mac) to make a fair conparison.
All of these models are implemented by Pytorch and ran on a single Tesla V100 GPU environment.
% Three test sets are used. 
As we can see, the prediction speed of Mac is around 5 times faster than Trans$_{E}$. 
Since the transition-based model employs a stack to store partially processed spans and a buffer to store unprocessed
tokens~\cite{dai-etal-2020-effective}, it is difficult to utilize GPU parallel computing to speed up the extraction process. 
In the official implementation, Trans$_{E}$ is restricted to process one token at a time, which means it is seriously inefficient and difficult to deploy in real development environment. 
By contrast, Mac is capable of handling data in batch mode because it is a single-stage sequence labeling model in essence.

\section{Conclusion}
In this paper, we reformulate discontinuous NER as the task of discovering maximal cliques in a segment graph, and propose a novel Mac architecture.
It decomposes the construction of segment graph as two independent 2-D grid tagging problems, and solves them jointly in one stage, addressing the exposure bias issue in previous studies.
Extensive experiments on three benchmark datasets show that Mac beats the previous SOTA method by as much as 3.5 pts in F1, while being 5 times faster.
Further analysis demonstrates the ability of our model in recognizing discontinuous and overlapping entity mentions.
In the future, we would like to explore similar formulation in other information extraction tasks, such as event extraction and nested NER.

% establishes a new state of art for discontinuous NER, 

% . We

% prove the superiority of Mac in both 

% % performs up to 11.4 times faster than previous 792 lattice-based method.

% propose a new discontinuous NER method – Mac, based on the inspiration that discontinuous NER 

% novel Iterative Grid Labeling archi- tecture, which models sequence labeling tasks with overlapping spans as a 2-D grid labeling problem.

% In this paper, we introduce the novel Mac architecture for discontinuous NER.
% It formulates 

% We propose a new discontinuous NER method – Mac, 

\section*{Acknowledgments}
We thank the reviewers for their insightful suggestions. 
This work is supported by the National Key Research and Development Program of China (Grant No.2017YFB0802804), the Guangdong Province Key
Area Research and Development Program of China (Grant No.2019B010137004), the Youth Innovation Promotion Association of Chinese Academy of Sciences (Grant No.2021153), and the Key Program of National Natural Science Foundation of China (Grant No.U1766215).

% The acknowledgments should go immediately before the references. Do not number the acknowledgments section.
% \textbf{Do not include this section when submitting your paper for review.}

\bibliographystyle{acl_natbib}
\bibliography{acl2021}

% \clearpage
% \appendix
% \input{ACL-2021-wyc/Sections/Appendix}
\end{document}